\documentclass[10pt,twocolumn,letterpaper]{article}

\usepackage{cvpr}              
\definecolor{cvprblue}{rgb}{0.21,0.49,0.74}
\usepackage[accsupp]{axessibility}
\usepackage[pagebackref,breaklinks,colorlinks,allcolors=cvprblue]{hyperref}
\usepackage{pifont}
\newcommand{\cmark}{\ding{51}}  
\newcommand{\xmark}{\ding{55}}  

\def\paperID{1518} 
\def\confName{CVPR}
\def\confYear{2026}

\title{GEM-TFL: Bridging Weak and Full Supervision for Forgery Localization through EM-Guided Decomposition and Temporal Refinement}

\author{
    Xiaodong Zhu$^{1}$ \quad Yuanming Zheng$^{1}$ \quad Suting Wang$^{1}$ \quad Junqi Yang$^{1}$\\ 
    Yuhong Yang$^{1,}$\thanks{Corresponding author.} \quad Weiping Tu$^{1}$ \quad Zhongyuan Wang$^{1}$ \\
    $^{1}$NERCMS, School of Computer Science, Wuhan University \\
    {\tt\small \{xiaodongzhu, ameixa, wangsuting, yangjq, yangyuhong, tuweiping\}@whu.edu.cn} \\
    {\tt\small wzy\_hope@163.com}
}

\begin{document}
\maketitle
\begin{abstract}
Temporal Forgery Localization (TFL) aims to precisely identify manipulated segments within videos or audio streams, providing interpretable evidence for multimedia forensics and security. While most existing TFL methods rely on dense frame-level labels in a fully supervised manner, Weakly Supervised TFL (WS-TFL) reduces labeling cost by learning only from binary video-level labels. However, current WS-TFL approaches suffer from mismatched training and inference objectives, limited supervision from binary labels, gradient blockage caused by non-differentiable top-$k$ aggregation, and the absence of explicit modeling of inter-proposal relationships. To address these issues, we propose GEM-TFL (\textbf{G}raph-based \textbf{EM}-powered \textbf{T}emporal \textbf{F}orgery \textbf{L}ocalization), a two-phase classification–regression framework that effectively bridges the supervision gap between training and inference. Built upon this foundation, (1) we enhance weak supervision by reformulating binary labels into multi-dimensional latent attributes through an EM-based optimization process; (2) we introduce a training-free temporal consistency refinement that realigns frame-level predictions for smoother temporal dynamics; and (3) we design a graph-based proposal refinement module that models temporal-semantic relationships among proposals for globally consistent confidence estimation. Extensive experiments on benchmark datasets demonstrate that GEM-TFL achieves more accurate and robust temporal forgery localization, substantially narrowing the gap with fully supervised methods.
\end{abstract}    
\section{Introduction}
\label{sec:intro}

The rapid progress of generative models has enabled the creation of realistic media, posing new challenges for content authenticity and digital forensics. Existing forgery detection methods~\cite{zhou2021joint, raza2023multimodaltrace, yu2023pvass, wang2023ftfdnet, smeu2025circumventing, oorloff2024avff, liu2023mcl, wu2025hola, nie2024frade} mainly focus on binary classification---determining whether a video is real or manipulated. In contrast, Temporal Forgery Localization (TFL)~\cite{cai2022you, cai2024av, zhang2023ummaformer, zhang2024mfms, perez2024vigo, cheng2025clformer, huang2025transhfc} aims to identify the precise temporal boundaries of forged segments, enabling interpretable and trustworthy forensic analysis. However, most TFL methods rely on dense frame-level labels, which are costly and difficult to scale, whereas clip-level authenticity labels (binary labels indicating whether an entire audio–visual sequence is real or fake) are much easier to obtain. This discrepancy motivates the research of Weakly Supervised TFL (WS-TFL), where models are trained using only clip-level labels but must temporally localize forgeries during inference.
\begin{figure}
    \centering
    \includegraphics[width=0.92\linewidth]{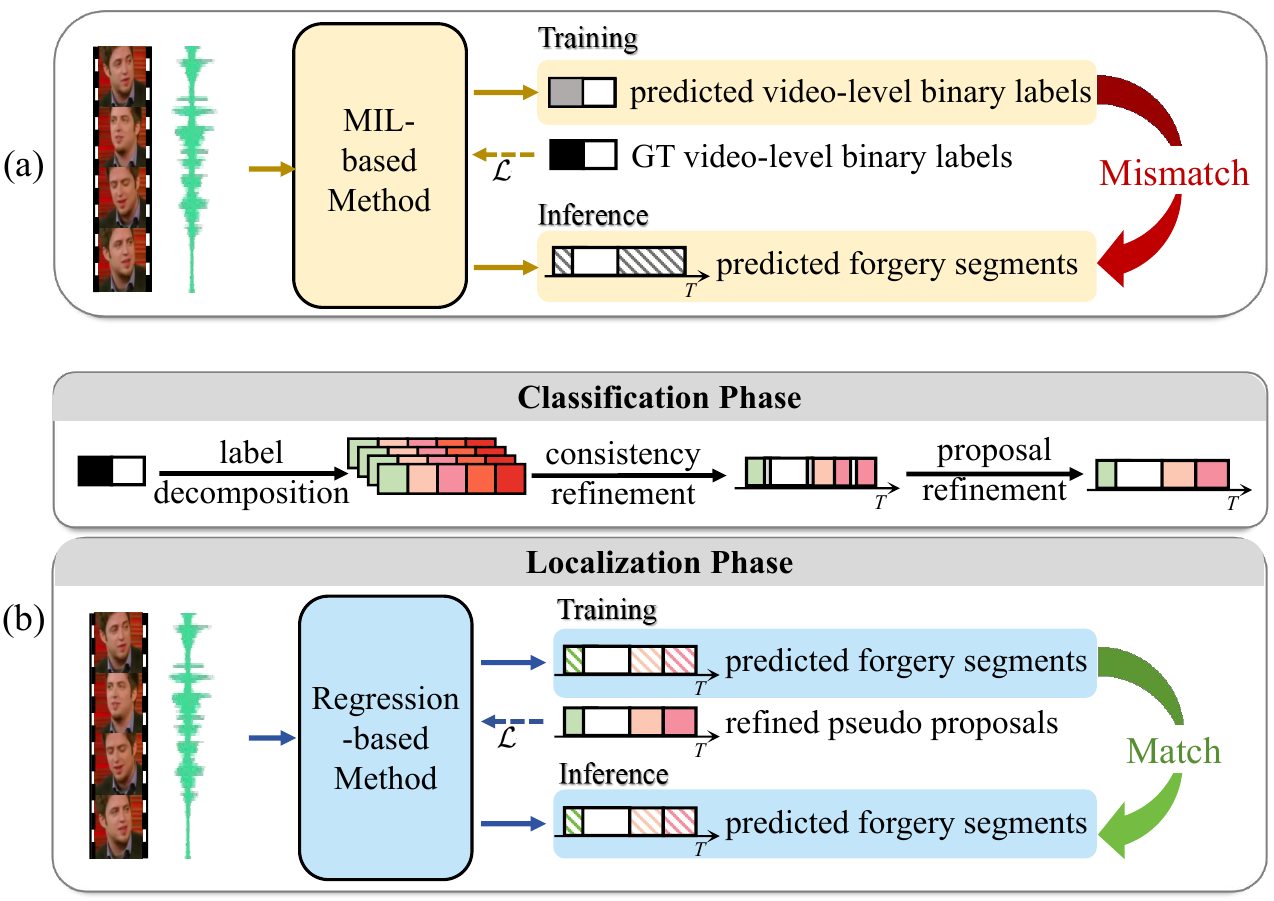}
    \caption{Comparison between prior and our WS-TFL pipelines. (a) Prior methods~\cite{xu2025multimodal, xu2025weaklysupervisedmultimodaltemporal}: trained with clip-level labels but required for temporal localization, leading to noisy proposals due to the training–inference mismatch. (b) Ours: a two-phase design aligning classification and regression to refine coarse predictions for precise boundary localization.}
    \label{fig:teaser}
\end{figure}

Several works (e.g., MDP~\cite{xu2025multimodal}, WMMT~\cite{xu2025weaklysupervisedmultimodaltemporal}) have explored WS-TFL under a Multiple Instance Learning (MIL) framework~\cite{maron1997framework} (shown in Figure~\ref{fig:teaser}(a)), similar to Temporal Action Detection (TAD)~\cite{wang2017untrimmednets, shou2018autoloc, zhang2021cola}. These methods aggregate frame-level activations and attention weights via a non-differentiable top-$k$ pooling operation to form clip-level predictions, and derive temporal proposals at inference through iterative thresholding and soft Non-Maximum Suppression (soft-NMS)~\cite{bodla2017soft}. However, the mismatch between training and inference objectives often leads to fragmented and unstable localization, especially for short and subtle forgeries.

Despite these limitations, several challenges remain unique to WS-TFL. 
1) \textbf{Limited supervision}. Unlike WS-TAD, which benefits from multi-class labels, WS-TFL relies on a single binary label, weakening semantic discrimination within the MIL framework. 
2) \textbf{Non-differentiable aggregation}. The top-$k$ pooling used to aggregate attention and activation sequences is inherently non-differentiable, blocking gradient flow and causing inconsistent temporal responses. While WS-TAD restores gradient propagation by jointly optimizing separate foreground and background losses~\cite{liu2025bridge}, this strategy fails in WS-TFL since binary supervision renders the two losses effectively identical, leaving the gradient blockage unresolved.
3) \textbf{Proposal fragmentation}. Conventional methods generate pseudo proposals by thresholding the activation sequence and computing Outer–Inner–Contrastive (OIC) scores~\cite{shou2018autoloc} for confidence estimation. This local approach ignores global dependencies among proposals, often fragmenting continuous forgeries into disjoint segments. Recent methods~\cite{liu2025bridge} attempt to fuse proposals into a unified global space weighted by OIC scores, but OIC computation is highly sensitive to the outer-region setting, introducing human bias and degrading proposal quality.

To overcome these limitations, we propose GEM-TFL, a two-phase classification–regression framework for weakly supervised temporal forgery localization (Figure~\ref{fig:teaser}(b)). In the first phase, a MIL-based classification branch generates a forgery activation sequence, from which pseudo proposals are derived to supervise a regression branch. This design bridges the objective gap between weakly and fully supervised settings. 
1) To enhance weak binary supervision, we introduce a Latent Attribute Decomposition (LAD) module that models latent forgery semantics through an Expectation–Maximization (EM) optimization~\cite{dempster1977maximum}. In the E-step, the posterior distribution over latent attributes is estimated—--assigning genuine samples to real class and distributing forged samples across multiple latent attributes based on model confidence. In the M-step, model parameters are updated to refine attribute separation and enrich semantic supervision. 
2) To address temporal inconsistencies caused by non-differentiable top-$k$ aggregation, we develop a Temporal Consistency Refinement (TCR) module that realigns frame-level predictions with clip-level attribute priors through a training-free constraint refinement, producing coherent temporal responses. 
3) A Graph-based Proposal Refinement (GPR) module further mitigates human bias by constructing a proposal relation graph combining temporal and semantic similarities and diffusing confidence across nodes for globally consistent proposal optimization.
In the Localization Phase (LP), a lightweight binary head provides auxiliary supervision via a binary classification loss, while the regression loss weight gradually increases during training to suppress noise from imperfect pseudo labels and ensure stable convergence.

To summarize, our contributions are as follows:
\begin{itemize}
    \item We propose GEM-TFL, a two-phase framework that bridges the gap between training and inference, substantially narrowing disparities between weakly and fully supervised TFL.  Overall, GEM-TFL achieves 8\% and 4\% absolute gains in average mAP on AV-Deepfake1M and LAV-DF, respectively.
    \item We propose an EM-based LAD module that transforms weak clip-level binary supervision into rich semantic attribute priors for improved representation learning.
    \item We introduce a training-free TCR module that realigns frame-level predictions with clip-level attribute priors through constraint-based refinement, producing smooth and stable temporal dynamics.
    \item We design a GPR module that constructs a proposal relation graph by integrating temporal and semantic similarities, diffusing confidence across neighbors to achieve globally consistent proposal optimization.
\end{itemize}

\section{Related Work}
\label{sec:related}

\begin{figure*}
    \centering
    \includegraphics[width=0.95\linewidth]{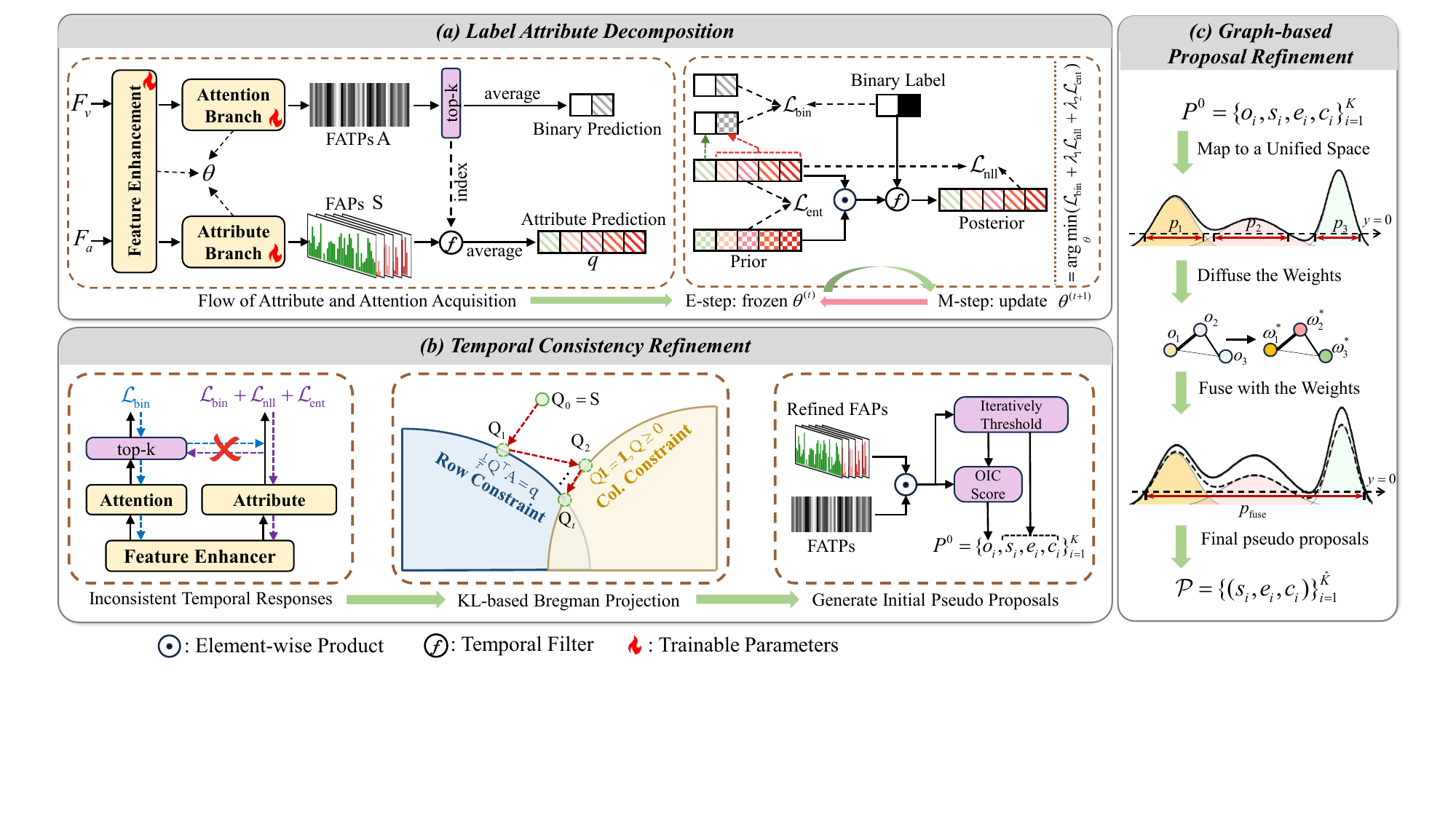}
    \caption{Overview of the classification phase. (a) Label Attribute Decomposition: The feature enhancement module aligns and fuses audio–visual features, after which the attention and attribute branches produce frame-level attention and attribute predictions optimized through EM to capture diverse forgery patterns. (b) Temporal Consistency Refinement: The non-differentiable top-$k$ operation blocks gradient flow between the attention and attribute branches, causing inconsistent temporal responses. To address this, frame-level attribute predictions are alternately projected onto the row constraint (attention-weighted alignment between frame- and clip-level predictions) and the column constraint (categorical distribution). The refined attribute predictions are then used to generate initial pseudo proposals. (c) Graph-based Proposal Refinement: Proposals are mapped into a unified space, where a proposal graph is constructed and confidence values are diffused to obtain fusion weights. These are integrated and thresholded at zero to produce the final pseudo proposals, merging fragmented proposals (e.g., $p_1$, $p_2$, $p_3$) into continuous ones (e.g., $p_{\text{fuse}}$) from a global and relational perspective.}
    \label{fig:overall}
\end{figure*}

\textbf{Fully-Supervised Deepfake Detection.} Early deepfake detection focused on multimodal scenario, leveraging audio–visual consistency via self-supervised~\cite{feng2023self, oorloff2024avff, lee2024multi, ilyas2023avfakenet} or contrastive objectives~\cite{zhou2021joint, raza2023multimodaltrace, yu2023pvass, katamneni2023mis, yu2023unified, liu2023audio}. Later works explored intra-modal relations~\cite{liu2023mcl, wang2024building} and differential mechanisms~\cite{smeu2025circumventing, koutlis2024dimodif, astrid2025audio}. With large-scale benchmarks such as LAV-DF~\cite{cai2023glitch} and AV-Deepfake1M~\cite{cai2024av}, research shifted to TFL, aiming to identify precise temporal boundaries of forged segments. Representative methods including BA-TFD~\cite{cai2022you}, UMMAFormer~\cite{zhang2023ummaformer}, MFMS~\cite{zhang2024mfms}, and RegQAV~\cite{zhu2025query} build on TAD frameworks including ActionFormer~\cite{zhang2022actionformer} and TriDet~\cite{shi2023tridet} with attention, query-based decoding, and multimodal fusion. Other works, such as Vigo~\cite{perez2024vigo}, TransHFC~\cite{huang2025transhfc}, and CLFormer~\cite{cheng2025clformer}, further enhance cross-modal robustness. Despite progress, these methods remain depend on dense frame-level labels, motivating weakly supervised alternatives.

\noindent\textbf{Weakly-supervised Video Localization.} Weakly supervised learning has shown promise in TAD, commonly formulated under the MIL paradigm~\cite{maron1997framework}. UntrimmedNet~\cite{wang2017untrimmednets} first introduced MIL for clip-level training, AutoLoc~\cite{shou2018autoloc} proposed an Outer–Inner–Contrastive (OIC) loss for snippet-level supervision, and CoLA~\cite{zhang2021cola} introduced snippet-level contrastive refinement. Later, PivoTAL~\cite{rizve2023pivotal} adopted a localization-by-localization paradigm, and PseudoFormer~\cite{liu2025bridge} improved pseudo label quality through the RickerFusion~\cite{gholamy2014ricker} strategy and label refinement.
Building on these advances, WS-TFL extends the MIL paradigm to forgery localization. Visual-based methods such as CoDL~\cite{li2025bi}, FuSTAL~\cite{feng2025full} and CPL~\cite{zhang2025cpl} target partially forged segments, while LOCO~\cite{wu2025weaklysupervisedaudiotemporalforgery} enhances audio-based localization through co-learning and self-supervision. For multimodal learning, MDP~\cite{xu2025multimodal} leverages cross-modal interactions and deviation-perceiving losses. WMMT~\cite{xu2025weaklysupervisedmultimodaltemporal} further introduces multimodal labels, enabling the model to predict modality-specific forgery types to enhance performance. However, this partially violates the weak supervision assumption of WS-TFL. Despite progress, WS-TFL still faces limited supervision, training–inference mismatch, and weak proposal reasoning, which we tackle in this work.


\section{Methodology}

\subsection{Problem Formulation}

Given a dataset of untrimmed talking-face videos, each sample $x=\{x^v,x^a\}$ contains temporally aligned video and audio streams. The WS-TFL aims to localize the timestamps of all forged segments in deepfake video, depending solely on the clip-level label $y\in\{0, 1\}$. The predicted forged segments are denoted as $\{(s_i,e_i,o_i)\}_{i=1}^M$, where $s_i$ and $e_i$ represent the start and end timestamps of the $i$-th forgery, and $o_i$ is the corresponding confidence score.

\subsection{Latent Attribute Decomposition (LAD)}

In WS-TFL, each video has a binary label indicating whether forgery exists, providing only limited semantic supervision. While the MIL framework can be applied, such sparse supervision hinders effective learning. To enrich semantics while maintaining weak supervision, we decouple the binary label into an $(m+1)$-dimensional latent attribute set $\mathcal{C} = \{0\} \cup \{1, \dots, m\}$, where $0$ denotes the real class and the remaining $m$ dimensions correspond to learnable latent forgery attributes. This formulation allows the model to capture diverse forgery patterns and enhances supervision granularity without requiring additional labels.

For the $i$-th sample $x_i = \{x_i^v, x_i^a\}$, we first extract frame-level visual features $F_v \in \mathbb{R}^{T_v \times C_v}$ and audio features $F_a \in \mathbb{R}^{T_a \times C_a}$ using pre-trained encoders. The two modalities are temporally and channel-wise aligned to produce a unified feature representation of dimension $T \times C$, facilitating joint audio–visual modeling. The aligned features are then refined through a Feature Enhancement Module consisting of Self-Attention (S-Attn), Cross-Attention (C-Attn), and Feed-Forward Network (FFN) layers, yielding audio-enhanced visual features and visual-enhanced audio features as follows:
\begin{equation}
    \begin{aligned}
\tilde{F}_v &= \text{FFN}_1(\text{C-Attn}_1(F_v, F_a) + \text{S-Attn}_1(F_v)),\\
\tilde{F}_a &= \text{FFN}_2(\text{C-Attn}_2(F_a, F_v) + \text{S-Attn}_2(F_a)),
\end{aligned}
\label{eq:feature_extraction}
\end{equation}

The enhanced features $\tilde{F}_v$ and $\tilde{F}_a$ are concatenated along the channel dimension to form a fused representation $F \in \mathbb{R}^{T \times 2C}$. The fused features are then passed into two branches: a forgery attention branch and a forgery attribute branch, producing Frame-level Attention Predictions (FATPs) $A$ and Frame-level Attribute Predictions (FAPs) $S$. Each branch employs an independent multilayer perceptron (MLP) for output generation:
\begin{equation}
    \begin{aligned}
A&=\operatorname{Sigmoid}(\operatorname{MLP}_1(F)),\\
S&=\operatorname{Softmax}(\operatorname{MLP}_2(F)),
\end{aligned}
\label{eq:FAP_SCP}
\end{equation}

The resulting $A \in \mathbb{R}^{T \times 1}$ denotes the forgery probability of each frame, while $S \in \mathbb{R}^{T \times (m+1)}$ represents the frame-level attribute predictions. The clip-level attribute prediction $q \in \mathbb{R}^{m+1}$ is then obtained by applying a top-$k$ aggregation~\cite{wang2017untrimmednets} over the FAPs guided by the attention scores:
\begin{equation}
    \begin{aligned}
&q=f_{agg}(S)=\operatorname{Softmax}\left(\frac{1}{k}\sum_{t\in\Omega}S_t\right),\\
&s.t.\quad \Omega=\arg \max_{\Omega}\sum_{t\in\Omega}A_t ,|\Omega|=k,
\end{aligned}
\label{eq:clip-level_attribution_prediction}
\end{equation}

Based on $q$, the clip-level forgery probability is computed as $\hat{y} = 1 - q_0$. In parallel, another clip-level prediction $\tilde{y}$ is obtained by applying the same temporal top-$k$ operation directly on the FAPs. The corresponding binary cross-entropy loss is defined as:
\begin{equation}
\begin{aligned}
\mathcal{L}_{\text{bin}}
= -\frac{1}{N}\sum_{i=1}^{N}\Big[&y_i\log \hat{y}_i + (1-y_i)\log(1-\hat{y}_i) + \\
&y_i\log \tilde{y}_i + (1-y_i)\log(1-\tilde{y}_i)
\Big],
\end{aligned}
\label{eq:BCE_loss}
\end{equation}

\noindent where $N$ is the number of samples. The binary cross-entropy loss enables the model to distinguish real from fake clips but not different forgery types. To address this, we apply the EM algorithm~\cite{dempster1977maximum} to optimize the latent attribute distribution, as shown in Figure~\ref{fig:overall}(a).

The EM algorithm consists of an E-Step and a M-Step. In the E-Step, given the model parameters $\theta^{(t)}$ at the $t$-th iteration and the binary label $y$, we compute the posterior distribution belonging to latent attribute $c$ as:
\begin{equation}
    \begin{aligned}
s_c&=\frac{\log \pi_c + \log q_c}{\tau},\\
P(c \mid x,y;\theta^{(t)})&=\begin{cases}
\mathds{1}[c=0], & y = 0, \\[6pt]
\displaystyle \frac{\mathds{1}[c\neq0]\exp(s_c)}{\sum_{j=1}^m \exp(s_j)}, \quad &y=1,
\end{cases}
\end{aligned}
\label{eq:EM_algorithm}
\end{equation}
Here, $\pi_c$ denotes the attribute prior, representing the distribution of the $c$-th attribute across the dataset, initialized as a uniform distribution $\pi_c = \tfrac{1}{m+1}$. The score $s_c$ is an unnormalized log value that combines prior and model evidence, and $\tau$ is the temperature coefficient (default set to $2$). $\mathds1[\cdot]$ denotes the indicator function. When $y = 0$, the posterior assigns all probability to $c = 0$; when $y = 1$, the probability mass is distributed over $c \neq 0$ via a softmax over the scores $s_{c=1}^m$.

In the M-step, we fix the posterior distribution and calculate the negative log-likelihood loss:
\begin{equation}
    \mathcal{L}_{\text{nll}}
= -\frac{1}{N}\sum_{i=1}^N \sum_{c=0}^{m} P(c \mid x_i, y_i; \theta^{(t)}) \log q_{i,c},
\label{eq: NLL_loss}
\end{equation}

To prevent the model from collapsing into few dominant attributes during training, we introduce an entropy-based regularization loss:
\begin{equation}
    \mathcal{L}_{\text{ent}}=-\sum_{c=1}^m\pi_c\log\left(\frac{1}{N}\sum_{i=1}^N q_{i,c}\right),
    \label{eq:ent_loss}
\end{equation}

Therefore, in the M-step, we update the model parameters by minimizing a combination of the binary cross-entropy loss, the negative log-likelihood loss, and the regularization loss:
\begin{equation}
    \theta^{(t+1)}=\arg\min_{\theta}(\mathcal{L}_{bin}+\lambda_1\mathcal{L}_{\text{nll}}+\lambda_2\mathcal{L}_{\text{ent}}),
    \label{eq:total_loss}
\end{equation}
Here, $\lambda_1$ and $\lambda_2$ denote the loss weights. 

The E-step and M-step alternate during training, forming a self-evolving label decomposition process that progressively refines both the latent attribute distribution and model parameters. After updating the model parameters, the class prior $\pi_c$ is dynamically updated via Exponential Moving Average (EMA):
\begin{equation}
    \pi_c^{(t+1)}=(1-\delta)\pi_c^{(t)}+\delta\cdot\frac{1}{N}\sum_{i=1}^N q_{i,c}.
    \label{eq:class_prior_update}
\end{equation}
where $\delta$ is the smoothing coefficient (default set to $0.0001$).

\subsection{Temporal Consistency Refinement (TCR)}

In the previous phase, the attribute and attention branches were connected through a non-differentiable top-$k$ aggregation (Figure~\ref{fig:overall}(b), left), which blocks gradient flow and leads to fragmented temporal responses under weak supervision.

To address this, we introduce a Temporal Consistency Refinement method (Figure~\ref{fig:overall}(b) middle) that realigns frame predictions with the clip-level distribution in a training-free manner. Let $S_t \in \mathbb{R}^{m+1}$ and $A_t \in [0,1]$ denote the frame-level attribute and attention predictions at time $t$. We seek an adjusted distribution $Q_t$ close to $S_t$ while aligning with the clip-level attribute prior $q$, formulated as the following KL-based Bregman projection problem:
\begin{equation}
    \begin{aligned} \min_Q\quad &\sum_{t=1}^T\operatorname{KL}(Q_t||S_t)=\sum_{t=1}^T\sum_{c=0}^mQ_{t,c}\log\frac{Q_{t,c}}{S_{t,c}},\\ &\operatorname{s.t.}\quad Q_{t,c}\ge 0,\sum_{c=0}^mQ_{t,c}=1,\forall t,\\ &\quad \quad \quad \frac{1}{T}\sum_{t=1}^TA_t Q_{t,c}=q_c,\forall c\in\mathcal{C}\setminus\{0\}, \end{aligned} 
    \label{eq:Bregman_optimize}
\end{equation}

The first constraint ensures that each $Q_t$ forms a valid categorical distribution, while the second enforces attention-weighted alignment between frame- and clip-level predictions. The optimization is solved using an Iterative Proportional Scaling (IPS) method~\cite{1940IPS}, which starts from $Q=S$ and alternately projects $Q$ onto the row and column constraint spaces until convergence. This process efficiently satisfies both constraints in a training-free manner. Detailed pseudocode is provided in Appendix~B.

After obtaining the optimized $Q$, we generate preliminary forgery proposals $P^{(0)} = \{(s_i, e_i, o_i, c_i)\}_{i=1}^K$ through iterative thresholding (Figure~\ref{fig:overall}(b), right), where $K$ is the number of proposals and $(s_i, e_i, o_i, c_i)$ denote the start time, end time, confidence, and attribute of each proposal $p_i$. Following AutoLoc~\cite{shou2018autoloc}, we compute the confidence score using the OIC score:
\begin{equation}
    o_i = \text{OIC}(p_i) = \text{avg}(p_i^{\text{inner}}) - \text{avg}(p_i^{\text{outer}}).
    \label{eq:OIC_score}
\end{equation}
where $p_i^{\text{inner}}$ and $p_i^{\text{outer}}$ represent the FAP values within and outside the proposal with a fixed ratio $\alpha$ (default $0.25$).

\begin{table*}[ht]
\centering
\setlength{\tabcolsep}{3pt} 
\begin{tabular}{c|l|c|cccccccc|ccccc}
\specialrule{1.2pt}{0pt}{0pt}
\multirow{2}{*}{\textbf{Supervision}} & \multirow{2}{*}{\textbf{Method}} & \multirow{2}{*}{\textbf{Modality}}  & \multicolumn{8}{c|}{\textbf{mAP@IoU(\%)}} & \multicolumn{5}{c}{\textbf{mAR@Proposals(\%)}} \\ 
&  &  & 0.1  & 0.2  & 0.3  & 0.4  & 0.5  & 0.6  & 0.7  & Avg. & 20 & 10 & 5 & 2 & Avg.  \\
\specialrule{1pt}{0pt}{0pt}
\multirow{4}{*}{Fully}       & AFormer {\tiny ECCV'22} \cite{zhang2022actionformer} & V & \cellcolor[HTML]{BDD9EF}{98.0} & \cellcolor[HTML]{BDD9EF}{97.7} & 97.3 & 96.8 & 96.3 & 95.5 & 94.6 & 96.6 & \cellcolor[HTML]{FFC2C1}{99.2}  & \cellcolor[HTML]{FFC2C1}{99.0}  & \cellcolor[HTML]{BDD9EF}{98.4}  & \cellcolor[HTML]{FFC2C1}{96.0}  & \cellcolor[HTML]{FFC2C1}{98.1}  \\
                             & TriDet {\tiny CVPR'23} \cite{shi2023tridet}    & V    & 95.0 & 94.7 & 94.4 & 93.8 & 93.2 & 92.2 & 90.7 & 93.4 & 97.1  & 96.9  & 96.3  & 93.7  & 96.0  \\
                             & UFormer {\tiny MM'23} \cite{zhang2023ummaformer}   & AV   & 97.7 & 97.6 & \cellcolor[HTML]{BDD9EF}{97.4} & \cellcolor[HTML]{BDD9EF}{97.1} & \cellcolor[HTML]{BDD9EF}{96.7} & \cellcolor[HTML]{BDD9EF}{96.0} & \cellcolor[HTML]{BDD9EF}{94.9} & \cellcolor[HTML]{BDD9EF}{96.8} & 98.6  & 98.5  & 98.2  & 95.1  & 97.6  \\
                             & MFMS {\tiny MM'24} \cite{zhang2024mfms}     & AV      & \cellcolor[HTML]{FFC2C1}{98.0} & \cellcolor[HTML]{FFC2C1}{97.9} & \cellcolor[HTML]{FFC2C1}{97.8} & \cellcolor[HTML]{FFC2C1}{97.6} & \cellcolor[HTML]{FFC2C1}{97.3} & \cellcolor[HTML]{FFC2C1}{96.7} & \cellcolor[HTML]{FFC2C1}{95.8} & \cellcolor[HTML]{FFC2C1}{97.3} & \cellcolor[HTML]{BDD9EF}{98.9}  & \cellcolor[HTML]{BDD9EF}{98.9}  & \cellcolor[HTML]{FFC2C1}{98.6}  & \cellcolor[HTML]{BDD9EF}{95.6}  & \cellcolor[HTML]{BDD9EF}{98.0}  \\ 
\hline
\multirow{7}{*}{Weakly}      & CoLA {\tiny CVPR'21} \cite{zhang2021cola}    & V      & 31.3 & 25.9 & 19.4 & 13.4 & 8.6  & 5.2  & 2.7  & 15.2 & 41.2  & 41.2  & 40.8  & 37.8  & 40.3  \\
                             & FuSTAL {\tiny TCSVT'25} \cite{feng2025full} & V  & 31.6  & 25.4  & 19.2  & 13.5   & 8.91  & 5.58  & 2.95  & 15.3  & 39.1  & 39.1  & 38.8  & 36.1  & 38.2  \\
                             & LOCO {\tiny IJCAI'25} \cite{wu2025weaklysupervisedaudiotemporalforgery}   & A     & 62.4 & 55.1 & 50.8 & 45.2 & 36.8 & 31.7 & 28.0 & 44.3 & 52.3  & 52.3  & 52.3  & 52.2  & 52.3  \\
                             & MDP {\tiny MM'25} \cite{xu2025multimodal}     & AV       & 76.5  & 75.9  & 71.2  & 62.0  & 50.8  & 44.1  & 39.7  & 60.0     & 69.2      & 66.7      & 60.7      & 51.7   & 62.1   \\
                             & PFormer {\tiny CVPR'25} \cite{liu2025bridge}  & V  & 87.1     & 77.5     & 73.2     & 64.2     & 57.7     & 50.9     & 42.4     & 64.7     & 81.5      & 80.3      & 80.1      & 78.5      & 80.1      \\
                             & WMMT {\tiny Arxiv'25} \cite{xu2025weaklysupervisedmultimodaltemporal}   & AV     & \cellcolor[HTML]{BDD9EF}{92.3} & \cellcolor[HTML]{BDD9EF}{86.7} & \cellcolor[HTML]{BDD9EF}{80.4} & \cellcolor[HTML]{BDD9EF}{74.7} & \cellcolor[HTML]{BDD9EF}{68.2} & \cellcolor[HTML]{BDD9EF}{60.9} & \cellcolor[HTML]{BDD9EF}{49.6} & \cellcolor[HTML]{BDD9EF}{73.3} & \cellcolor[HTML]{BDD9EF}{85.6}  & \cellcolor[HTML]{BDD9EF}{85.6}  & \cellcolor[HTML]{BDD9EF}{85.6}  & \cellcolor[HTML]{FFC2C1}{85.3}  & \cellcolor[HTML]{BDD9EF}{85.5}  \\
\cline{2-16}\addlinespace[0.5pt]
                             & GEM-TFL(Ours)  & AV   & \cellcolor[HTML]{FFC2C1}{93.7}     & \cellcolor[HTML]{FFC2C1}{89.1}     & \cellcolor[HTML]{FFC2C1}{85.6}     & \cellcolor[HTML]{FFC2C1}{79.3}     & \cellcolor[HTML]{FFC2C1}{72.5}     & \cellcolor[HTML]{FFC2C1}{66.2}     & \cellcolor[HTML]{FFC2C1}{57.1}     & \cellcolor[HTML]{FFC2C1}{77.6}     & \cellcolor[HTML]{FFC2C1}{88.2}      & \cellcolor[HTML]{FFC2C1}{87.5}      & \cellcolor[HTML]{FFC2C1}{86.1}      & \cellcolor[HTML]{BDD9EF}{84.3}  & \cellcolor[HTML]{FFC2C1}{86.5}   \\
\specialrule{1.2pt}{0pt}{0pt}
\end{tabular}
\caption{Temporal forgery localization results of fully and weakly supervised methods on LAV-DF~\cite{cai2023glitch}. 'UFormer' and 'PFormer' denote UMMAFormer~\cite{zhang2023ummaformer} and PseudoFormer~\cite{liu2025bridge}. 'V', 'A', and 'AV' indicate visual, audio, and audio–visual modalities. Fully supervised methods use forged-segment timestamps as supervision, while weakly supervised ones use binary clip-level labels, except WMMT~\cite{xu2025weaklysupervisedmultimodaltemporal}, which adopts a quadruple signal (both real, both fake, visual fake only, and audio fake only). The best mAP/mAR are highlighted in \colorbox[HTML]{FFC2C1}{red}, and the second-best in \colorbox[HTML]{BDD9EF}{blue}.}
\label{table1}
\end{table*}

\subsection{Graph-based Proposal Refinement (GPR)}

\begin{figure}[ht]
    \centering
    \includegraphics[width=\linewidth]{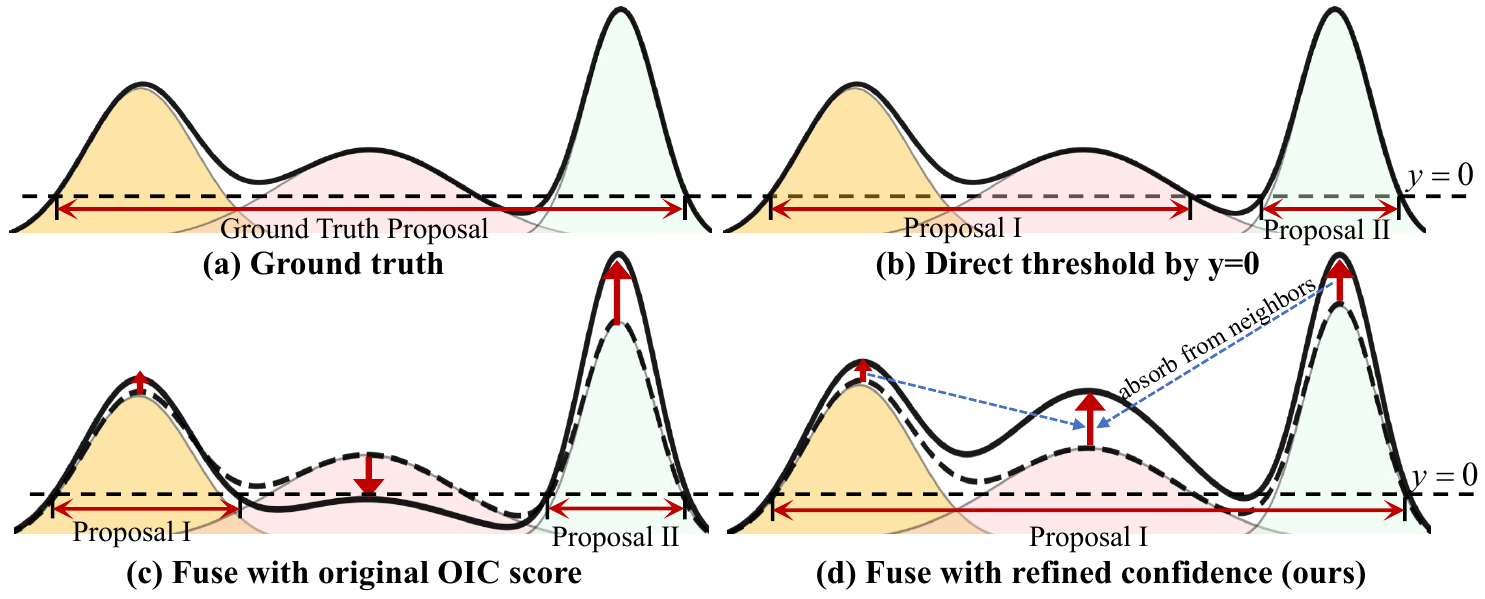}
    \caption{Comparison between prior and our WS-TFL pipelines.}
    \label{fig:graph}
\end{figure}

Following PseudoFormer~\cite{liu2025bridge}, we map all preliminary proposals $P^{(0)}$ into a unified global space using the Ricker wavelet~\cite{gholamy2014ricker} to enhance global perception:
\begin{equation}
    \phi_i(t)=\frac{2}{\sqrt{3\sigma_i}\pi^{\frac{1}{4}}}\left(1-\left(\frac{t-m_i}{\sigma_i}\right)^2\right)e^{-\frac{(t-m_i)^2}{2\sigma^2_i}},
    \label{eq:Gaussian_kernel_mapping}
\end{equation}
where $\sigma_i=\frac{e_i-s_i}{2}$ and $m_i=\frac{s_i+e_i}{2}$ denote the proposal length and center. In this setting, proposals are fused in a global space using OIC-based confidences~\cite{shou2018autoloc}. However, the OIC magnitude is highly sensitive to the outer-region parameter $\alpha$: larger $\alpha$ values incorporate more external information, leading to smaller OIC scores and inconsistent fusion behavior. Compare Figure~\ref{fig:graph}(b) with Figure~\ref{fig:graph}(c), directly using these OIC scores as fusion weights introduces human bias, producing unstable or distorted global representations.

To mitigate this bias, we refine proposal confidence based on inter-proposal relationships. As shown in Figure~\ref{fig:overall}(c), we construct an undirected graph $G=(V,E)$, where each node represents a proposal $p_i \in P^{(0)}$. The edge weight $e_{ij}$ between $p_i$ and $p_j$ combines temporal similarity $d_{ij}^p$ (measured by DIoU~\cite{rezatofighi2019diou}) and semantic similarity $d_{ij}^s$, defined as:
\begin{equation}
\begin{aligned}
d_{ij}^s=\begin{cases}1, &c_i=c_j,\\
1/m, &c_i\ne c_j,\end{cases}
\end{aligned}
\label{eq:semantic_similarity}
\end{equation}

The overall edge weight is defined as $e_{ij} = d_{ij}^p + 0.5\times d_{ij}^s$. From the adjacency matrix, we derive the row-normalized transition matrix $\mathcal{T}_{i,j} = e_{ij} / \sum_j e_{ij}$. The initial kernel weights are set to the proposal confidences, $\omega_i^{(0)} = o_i$, and iteratively diffused across the graph to propagate supportive evidence among neighbors: 
\begin{equation}
    \omega^{(t+1)}=\beta \mathcal{T}\omega^{(t)}+(1-\beta) \omega^{(0)},\beta=0.7
    \label{eq:supportive_evidence}
\end{equation}

Since $\beta \in (0,1)$, this recurrence admits a closed-form solution(see the proof in Appendix C): $\omega^* = (1 - \beta)(I - \beta \mathcal{T})^{-1}\omega^{(0)}$. Finally, the global representation is obtained as:
\begin{equation}
     \Phi^l=\sum_{i=1}^K \omega_i^{*} \phi_i\mathds{1}[c_i=l].
     \label{eq:refined confidence}
\end{equation}
where $\Phi \in \mathbb{R}^{T\times m}$. The final pseudo labels $\mathcal{P} = \{(\hat{s}_i, \hat{e}_i, \hat{c}_i)\}_{i=1}^{\hat{K}}$ are derived by thresholding the activated regions of $\Phi$ at $y = 0$. This graph-based diffusion refines proposal confidences, yielding more reliable and coherent temporal boundaries (Figure~\ref{fig:graph}(d)).

\subsection{Localization Phase (LP)}

In the second phase, a regression branch (e.g., UMMAFormer~\cite{zhang2023ummaformer}, TriDet~\cite{shi2023tridet}) is trained under the supervision of pseudo proposals generated in the first phase, enabling precise boundary localization consistent with fully supervised TFL. The regression branch predicts the clip-level forgery probability $\hat{y}$ and regresses temporal proposals $\hat{\mathcal{P}}$. To mitigate noise from imperfect pseudo labels, a binary classification head (two-layer MLP) is attached to the regression features. Its logits are used to compute a binary cross-entropy loss $\mathcal{L}_{\text{bce}}(\hat{y}, y)$, which is jointly optimized with the main regression loss $\mathcal{L}_{\text{main}}(\hat{\mathcal{P}}, \mathcal{P})$:
\begin{equation} 
\mathcal{L}_{\text{total}} = \mathcal{L}_{\text{bce}}(\hat{y}, y) + \gamma \cdot \mathcal{L}_{\text{main}}(\hat{\mathcal{P}}, \mathcal{P}).
\label{eq:final} 
\end{equation}
where $\gamma$ increases linearly from 0.5 to 1.0 during training. During inference, only the regression branch is used, followed by soft-NMS~\cite{bodla2017soft} to produce the final predictions, effectively bridging the supervision gap between training and inference.
\section{Experiments}

\begin{table*}[ht]
\centering
\setlength{\tabcolsep}{3pt} 
\begin{tabular}{c|l|c|cccccccc|ccccc}
\specialrule{1.2pt}{0pt}{0pt}
\multirow{2}{*}{\textbf{Supervision}} & \multirow{2}{*}{\textbf{Method}} & \multirow{2}{*}{\textbf{Modality}}  & \multicolumn{8}{c|}{\textbf{mAP@IoU(\%)}} & \multicolumn{5}{c}{\textbf{mAR@Proposals(\%)}} \\ 
&  &  & 0.1  & 0.2  & 0.3  & 0.4  & 0.5  & 0.6  & 0.7  & Avg. & 20 & 10 & 5 & 2 & Avg.  \\
\specialrule{1pt}{0pt}{0pt}
\multirow{4}{*}{Fully}       & AFormer {\tiny ECCV'22} \cite{zhang2022actionformer} & V & 67.3 & 67.3 & 67.2 & 67.2 & 67.0 & 66.7 & 65.5 & 66.9 & 83.1  & 82.9  & 82.6  & 78.7  & 81.8  \\
                             & TriDet {\tiny CVPR'23} \cite{shi2023tridet}   & V     & 55.7 & 55.6 & 55.4 & 55.1 & 54.7 & 53.8 & 51.7 & 54.6 & 74.9 & 74.2  & 72.9  & 68.0  & 72.5  \\
                             & UFormer {\tiny MM'23} \cite{zhang2023ummaformer}  & AV    & \cellcolor[HTML]{BDD9EF}{91.8} & \cellcolor[HTML]{BDD9EF}{91.7} & \cellcolor[HTML]{BDD9EF}{91.5} & \cellcolor[HTML]{BDD9EF}{91.3} & \cellcolor[HTML]{BDD9EF}{91.0} & \cellcolor[HTML]{BDD9EF}{90.4} & \cellcolor[HTML]{BDD9EF}{89.0} & \cellcolor[HTML]{BDD9EF}{90.9} & \cellcolor[HTML]{BDD9EF}{95.0}  & \cellcolor[HTML]{BDD9EF}{94.6}  & \cellcolor[HTML]{BDD9EF}{93.9}  & \cellcolor[HTML]{BDD9EF}{89.5}  & \cellcolor[HTML]{BDD9EF}{93.3}  \\
                             & MFMS {\tiny MM'24} \cite{zhang2024mfms}     & AV       & \cellcolor[HTML]{FFC2C1}{94.7} & \cellcolor[HTML]{FFC2C1}{94.6} & \cellcolor[HTML]{FFC2C1}{94.6} & \cellcolor[HTML]{FFC2C1}{94.5} & \cellcolor[HTML]{FFC2C1}{94.3} & \cellcolor[HTML]{FFC2C1}{93.9} & \cellcolor[HTML]{FFC2C1}{92.5} & \cellcolor[HTML]{FFC2C1}{94.1} &\cellcolor[HTML]{FFC2C1}{96.7}  & \cellcolor[HTML]{FFC2C1}{96.4}  & \cellcolor[HTML]{FFC2C1}{96.0}  & \cellcolor[HTML]{FFC2C1}{92.0}  & \cellcolor[HTML]{FFC2C1}{95.3}  \\ 
\hline
\multirow{7}{*}{Weakly}      & CoLA {\tiny CVPR'21} \cite{zhang2021cola}    & V      & 3.22 & 1.09 & 0.39 & 0.14 & 0.05  & 0.02  & 0.01  & 0.70  & 20.7  & 20.0  & 16.1  & 8.17  & 16.2  \\
                             & FuSTAL {\tiny TCSVT'25} \cite{feng2025full} & V   & 3.03  & 1.02  & 0.40  & 0.15  & 0.05  & 0.02  & 0.01  & 0.67  & 19.7  & 18.6  & 14.6  & 7.08  & 15.0  \\
                             & LOCO {\tiny IJCAI'25} \cite{wu2025weaklysupervisedaudiotemporalforgery}   & A      & 1.25 & 0.30 & 0.10 & 0.03 & 0.01 & 0.00 & 0.00 & 0.24   & 10.4  & 10.1  & 8.48  & 3.99  & 8.25  \\
                             & MDP {\tiny MM'25} \cite{xu2025multimodal}     & AV        & 5.35  & 3.33  &  1.83 & 0.86  & 0.36  & 0.14  & 0.05  & 3.68 &   39.5  & 36.9      & 28.6      & 19.8      & 31.2      \\
                             & PFormer {\tiny CVPR'25} \cite{liu2025bridge}  & V  & 38.4     & 34.2     & 30.7     & 24.6     & 17.1     & 14.5     & 5.53     & 23.6     &   \cellcolor[HTML]{BDD9EF}{55.7}      & \cellcolor[HTML]{BDD9EF}{52.9}      & 51.3      & \cellcolor[HTML]{BDD9EF}{50.4}      & \cellcolor[HTML]{BDD9EF}{52.6}      \\
                             & WMMT {\tiny Arxiv'25}  \cite{xu2025weaklysupervisedmultimodaltemporal}   & AV       & \cellcolor[HTML]{BDD9EF}{54.5} & \cellcolor[HTML]{BDD9EF}{48.4} & \cellcolor[HTML]{BDD9EF}{43.6} & \cellcolor[HTML]{BDD9EF}{38.5} & \cellcolor[HTML]{BDD9EF}{29.3} & \cellcolor[HTML]{BDD9EF}{18.8} & \cellcolor[HTML]{BDD9EF}{7.33} & \cellcolor[HTML]{BDD9EF}{34.3} &   52.8  & 52.7  & \cellcolor[HTML]{BDD9EF}{52.2}  & 50.2  & 52.0  \\
\cline{2-16}\addlinespace[0.5pt]
                             & GEM-TFL(Ours) & AV   & \cellcolor[HTML]{FFC2C1}{65.8}     & \cellcolor[HTML]{FFC2C1}{58.1}     & \cellcolor[HTML]{FFC2C1}{52.3}     & \cellcolor[HTML]{FFC2C1}{44.9}     & \cellcolor[HTML]{FFC2C1}{38.9}     & \cellcolor[HTML]{FFC2C1}{26.4}     & \cellcolor[HTML]{FFC2C1}{12.2}     & \cellcolor[HTML]{FFC2C1}{42.7}       & \cellcolor[HTML]{FFC2C1}{63.7}      & \cellcolor[HTML]{FFC2C1}{60.2}      & \cellcolor[HTML]{FFC2C1}{58.9}      & \cellcolor[HTML]{FFC2C1}{56.4}      & \cellcolor[HTML]{FFC2C1}{59.8}      \\
\Xhline{1.2pt}
\end{tabular}
\caption{Temporal forgery localization results of fully and weakly supervised methods on the AV-Deepfake1M \cite{cai2024av} benchmark.}
\label{table2}
\end{table*}

\subsection{Experimental Setup}


\textbf{Datasets.} We evaluate our method on two challenging multimodal Deepfake temporal forgery datasets: LAV-DF~\cite{cai2023glitch} and AV-Deepfake1M~\cite{cai2024av}. LAV-DF is a content-centric audio–visual dataset specifically designed for temporal localization, containing 136,304 videos across 153 identities. It includes only replacement-type manipulations, with forged segments averaging 0.65 s in videos of 8.58 s. AV-Deepfake1M is a large-scale multimodal dataset comprising 1,886 h of videos from 2,068 identities. It introduces fine-grained manipulations---including replacement, insertion, and deletion---for both audio and video, combining text from large language models with modern synthesis methods. Compared with LAV-DF, AV-Deepfake1M employs more advanced audio and visual synthesis techniques and offers larger scale, richer modalities, and a lower forgery ratio, making it a more challenging benchmark for weakly supervised TFL. More details are provided in Appendix A.

\noindent\textbf{Metrics.} We adopt Average Precision (AP) and Average Recall (AR) as our evaluation metrics, following the common settings~\cite{xu2025multimodal, xu2025weaklysupervisedmultimodaltemporal, liu2025bridge}. For both LAVDF and AV-Deepfake1M, the IoU thresholds of mAP are set as [0.1:0.1:0.7], and the number of proposals is set as 20, 10, 5, and 2, respectively.

\noindent\textbf{Baselines.} To validate the effectiveness of our method, we conduct comparisons with both fully supervised and weakly supervised localization approaches. For the fully supervised localization setting, we select ActionFormer~\cite{zhang2022actionformer}, TriDet~\cite{shi2023tridet}, UMMAFormer~\cite{zhang2023ummaformer}, and MFMS~\cite{zhang2024mfms} as baselines. Among them, ActionFormer and TriDet are representative action detection models, while UMMAFormer and MFMS are effective audio-visual forgery localization models. For the weakly supervised localization setting, we compare with three WS-TAD methods (CoLA~\cite{zhang2021cola}, FuSTAL~\cite{feng2025full}, PseudoFormer~\cite{liu2025bridge}), one audio WS-TFL method (LOCO~\cite{wu2025weaklysupervisedaudiotemporalforgery}), and two audio-visual WS-TFL methods (MDP~\cite{xu2025multimodal} and WMMT~\cite{xu2025weaklysupervisedmultimodaltemporal}). It is important to note that the weakly supervised methods rely solely on clip-level binary labels for supervision during training, except for WMMT, which utilizes multimodal labels represented as a quadruple: (both real, both fake, visual fake only, audio fake only).

\noindent\textbf{Implementation Details.} We employ ResNet-50~\cite{he2016deep} and Wav2Vec 2.0~\cite{baevski2020wav2vec} as visual and audio feature extractors, respectively. The unified feature dimension is 512, and the number of latent attributes $m$ is 3. During the M-step, loss weights $\lambda_1$ and $\lambda_2$ are 0.8 and 0.5. UMMAFormer is used as the regression model. Both the calssification and localization models are trained for 50 epochs on LAV-DF and AV-Deepfake1M with a batch size of 128, using AdamW~\cite{loshchilov2017decoupled} and a cosine learning rate schedule with a 10-epoch warm-up. The learning rate is set to 2e-4. All experiments are conducted on eight NVIDIA RTX 3090 GPUs.

\subsection{Performance Comparison}

In this section, we conduct a comparative evaluation of our approach against state-of-the-art fully supervised and weakly supervised TFL methods on the LAV-DF and AV-Deepfake1M datasets. For a fair comparison, all baseline methods were retrained on the same pre-trained features according to their official open-source implementations.

\textbf{LAVDF Dataset.} As shown in Table~\ref{table1}, our method achieves the best weakly supervised localization results in both mAP and mAR. Compared with existing WS-TAD methods, it surpasses the strongest baseline, PseudoFormer, by 12.7\% in average mAP and 6.4\% in average mAR. Against WS-TFL methods, it further outperforms the best-performing WMMT by 4.3\% and 1.0\% in average mAP and mAR, respectively. Moreover, our method markedly narrows the gap with fully supervised models on mAP@\{0.1, 0.2, 0.3\} and mAR@\{20, 10, 5\}. Although weakly supervised methods typically experience sharp mAP drops at higher IoU thresholds due to the lack of precise boundaries, our model maintains over 50\% mAP at IoU 0.7, demonstrating strong robustness and accurate boundary localization.

\textbf{AV-Deepfake1M Dataset.} As shown in Table~\ref{table2}, our method still achieves the best weakly supervised localization performance in terms of both mAP and mAR. Due to the larger scale and more diverse forgery methods in AV-Deepfake1M, as well as the more challenging boundary localization, both fully and weakly supervised methods show a noticeable performance drop on this dataset. CoLA, FuSTAL, and LOCO all achieve an average mAP of less than 1\%, indicating that these methods struggle to adapt to the complexity of AV-Deepfake1M. Although MDP is a multi-modal WS-TFL method, its performance also experiences a significant decline. PseudoFormer still performs reasonably well on AV-Deepfake1M, primarily due to its two-phase architecture. WMMT also achieves decent performance, mainly due to its use of multimodal labels, where the quadruple labels provide far richer semantic information than binary labels, compensating for the lack of precise boundary labels. Our method surpasses all other methods by 8.4\% in average mAP, significantly narrowing the gap with fully supervised methods. The success of our approach can be attributed to a combination of WMMT and PseudoFormer. On one hand, we significantly enrich the semantic content of the supervision through label disentangling; on the other hand, we achieve precise localization through pseudo-label optimization and a two-phase process.

\subsection{Ablation Study} 

\begin{table}[ht]
\centering
\setlength{\tabcolsep}{3.5pt} 
\begin{tabular}{c|cccc|ccc}
\specialrule{1.2pt}{0pt}{0pt}
\textbf{\#} & \textbf{LAD}                   & \textbf{TCR}                   & \textbf{GPR}                   & \textbf{LP}                    & \begin{tabular}[c]{@{}c@{}}\textbf{mAP}\\ (0.1:0.7)\end{tabular} & \multicolumn{1}{c}{\begin{tabular}[c]{@{}c@{}}\textbf{mAR}\\ (0.1:0.7)\end{tabular}} & \begin{tabular}[c]{@{}c@{}}\textbf{Param}\\ {[}M{]}\end{tabular} \\
\specialrule{1pt}{0pt}{0pt}
1  & \xmark & \xmark & \xmark & \xmark & 5.12 & 35.6  & 17.1                                                                           \\
2  & \cmark & \xmark & \xmark & \xmark & 23.8{\tiny $\uparrow$18.7} & 44.3{\tiny $\uparrow$8.7}   & 17.2                                                                          \\
3  & \cmark & \cmark & \xmark & \xmark & 27.3{\tiny $\uparrow$22.2} & 46.7{\tiny $\uparrow$11.1}    & 17.2                                                                            \\
4  & \cmark & \cmark & \cmark & \xmark & 31.9{\tiny $\uparrow$26.8} & 51.2{\tiny $\uparrow$15.6}     & 17.2                                                                           \\ \hline
5  & \cmark & \cmark & \cmark & \cmark & 42.7{\tiny $\uparrow$37.6} & 59.8{\tiny $\uparrow$24.2}     & 62.2                                                                           \\
\specialrule{1.2pt}{0pt}{0pt}
\end{tabular}
\caption{Ablation study on the components. Performance gains relative to the base model (row \#1) are indicated as subscripts.}
\label{table3}
\end{table}

\noindent \textbf{Effectiveness of the Components}. We conduct ablation studies on AV-Deepfake1M to evaluate the contribution of each component (Table~\ref{table3}). Without any modules, the base model reduces to a MIL-based WS-TFL similar to MDP~\cite{xu2025multimodal}, where an attention branch slightly improves mAP (5.12\% vs. 3.68\%). Adding the LAD module, which semantically decouples binary supervision, yields large gains (+18.7\%, +8.7\%), showing the benefit of richer semantics. The TCR module aligns attribute and attention predictions in a training-free manner, alleviating fragmented localization and further improving performance (+3.5\%, +2.3\%). The GPR module models inter-proposal relations via graph reasoning, producing more stable boundaries (+4.6\%, +4.5\%). These three modules introduce negligible parameter overhead compared with the base model. The LP module adds 45M parameters (from UMMAFormer) but significantly boosts performance (+10.8\%, +8.6\%), demonstrating an effective trade-off between complexity and accuracy. Overall, LAD and LP contribute most to localization performance, consistent with WMMT~\cite{xu2025weaklysupervisedmultimodaltemporal} and PseudoFormer~\cite{liu2025bridge}, further verifying that enriched supervision effectively narrows the gap with fully supervised methods.

\begin{figure}[ht]
    \centering
    \includegraphics[width=\linewidth]{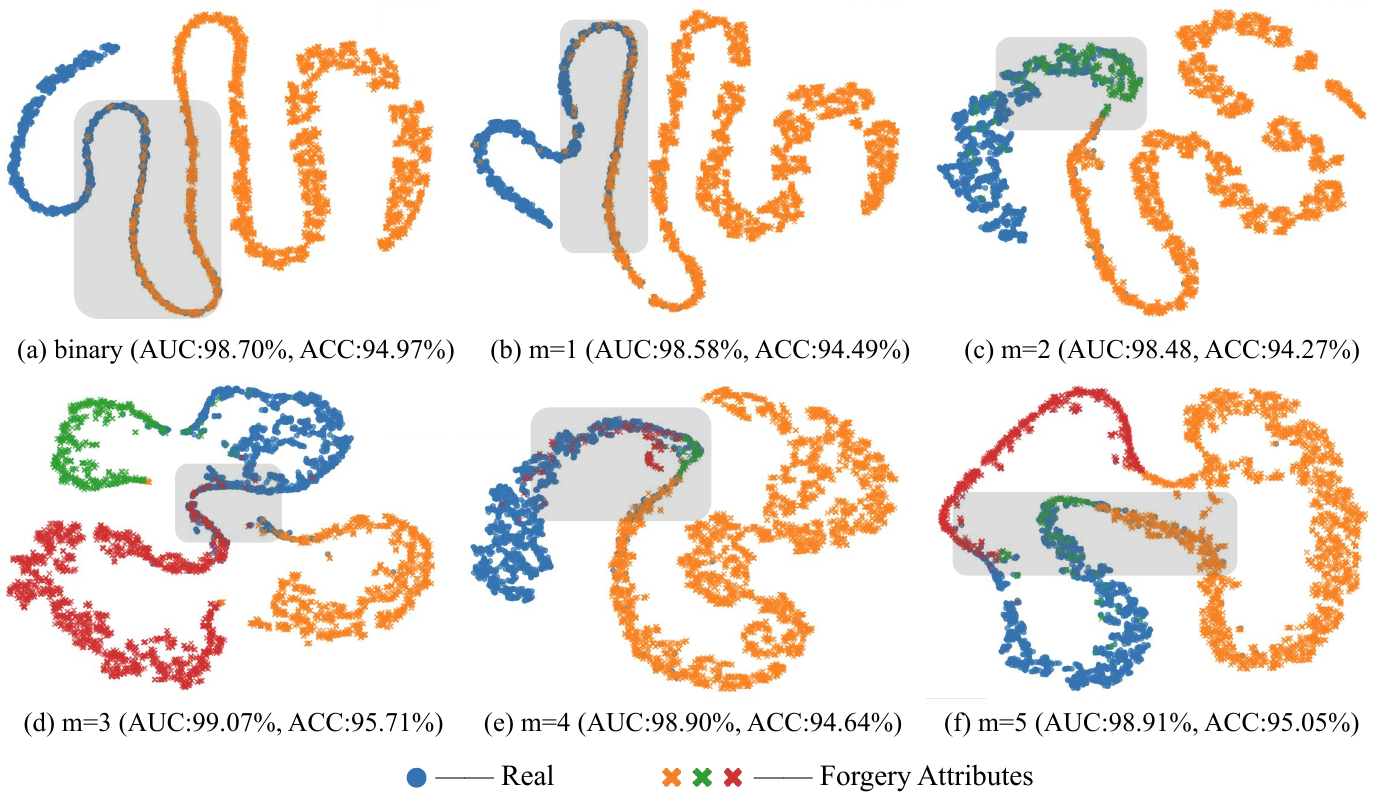}
    \caption{T-SNE visualizations of classification features under different latent forgery attribute d $m$. Overlaps between attributes are highlighted in gray. “binary” denotes the setting without EM optimization, where the model directly performs binary classification.}
    \label{fig:tse}
\end{figure}

\noindent \textbf{Different choices of $m$}. We conduct an ablation study on the choice of $m$. To visualize the effect of latent attribute decomposition, we perform t-SNE on the classification features and report binary classification Accuracy (ACC) and Area under the Curve (AUC). As shown in Figure~\ref{fig:tse}, when $m<3$, attribute clusters overlap heavily, indicating incomplete disentanglement; when $m>3$, the features collapse. The best disentanglement and classification performance occur at $m=3$, even surpassing direct binary classification, underscoring the model’s ability to expand the semantic space and enable more nuanced discrimination, thereby improving overall performance. We posit that $m=3$ is optimal because the latent attributes naturally align with three modality-level forgery patterns: audio-only, visual-only, and joint audio-visual. Under weak supervision, these modality-driven signals are inherently more discriminative and stable to learn than diverse specific forgery methods.

\subsection{Generalization Study} 

\begin{table}[ht]
\centering
\setlength{\tabcolsep}{3pt}
\begin{tabular}{l|cccccccc}
\specialrule{1.2pt}{0pt}{0pt}
\multirow{2}{*}{\textbf{Method}} & 
\multicolumn{8}{c}{\textbf{mAP@IoU(\%)}} \\ 
& 0.1  & 0.2  & 0.3  & 0.4  & 0.5  & 0.6  & 0.7  & Avg.  \\
\specialrule{1.2pt}{0pt}{0pt}
 CoLA        & 1.35 & 0.13 & 0.03 & 0.01 & 0.01 & 0.01 & 0.01 & 0.22 \\
 LOCO        & 0.27 & 0.04 & 0.01 & 0.00 & 0.00 & 0.00 & 0.00 & 0.05 \\
 MDP         & 15.5 & 9.10 & 7.76 & 4.01 & 2.33 & 1.83 & 0.42 & 5.85 \\
 PFormer     & \cellcolor[HTML]{BDD9EF}{26.3}     & \cellcolor[HTML]{BDD9EF}{22.7}     & \cellcolor[HTML]{BDD9EF}{16.6}     & 12.7     & 9.28     & 6.44     & 2.15     & \cellcolor[HTML]{BDD9EF}{13.7} \\
 WMMT        & 21.7 & 18.5 & 15.7 & \cellcolor[HTML]{BDD9EF}{13.1} & \cellcolor[HTML]{BDD9EF}{10.2} & \cellcolor[HTML]{BDD9EF}{7.39} & \cellcolor[HTML]{BDD9EF}{4.45} & 13.0\\
\cline{1-9}\addlinespace[0.5pt]
GEM-TFL   & \cellcolor[HTML]{FFC2C1}{30.5}     & \cellcolor[HTML]{FFC2C1}{23.9}     & \cellcolor[HTML]{FFC2C1}{20.4}     & \cellcolor[HTML]{FFC2C1}{16.8}     & \cellcolor[HTML]{FFC2C1}{12.4}     & \cellcolor[HTML]{FFC2C1}{9.89}     & \cellcolor[HTML]{FFC2C1}{5.78}     & \cellcolor[HTML]{FFC2C1}{17.1} \\
\specialrule{1.2pt}{0pt}{0pt}
\end{tabular}
\caption{Cross-dataset generalization performance. Models are trained on AV-Deepfake1M~\cite{cai2024av} and tested on LAV-DF~\cite{cai2023glitch}.}
\label{table4}
\end{table}

In this section, we evaluate the cross-dataset generalization of our method. Since AV-Deepfake1M is considerably more complex than LAV-DF, training on LAV-DF and testing on AV-Deepfake1M results in nearly all WS-TFL methods yielding 0.00 mAP@0.1. Thus, we only report results for training on AV-Deepfake1M and testing on LAV-DF (Table~\ref{table4}). Our method achieves the best cross-dataset generalization among all weakly supervised approaches, surpassing PseudoFormer by 3.4\% in average mAP. Notably, PseudoFormer performs better at low IoU thresholds but falls behind WMMT as IoU increases, underscoring the importance of semantically rich supervision for precise forgery localization and validating the effectiveness of our latent attribute decomposition module.



\section{Conclusion}


We propose a two-phase weakly supervised temporal forgery localization framework that bridges the gap between training and inference. Latent attribute decomposition and localization-phase regression enhance boundary precision and robustness, while temporal consistency refinement and graph-based proposal refinement improve temporal smoothness and global coherence. Together, these components enrich weak supervision with semantic, temporal, and structural cues, advancing weakly supervised TFL toward fully supervised performance. Although our method achieves notable gains over weakly supervised baselines (+8\% and +4\% mAP on AV-Deepfake1M and LAV-DF), a clear gap remains to fully supervised methods. Future work will explore leveraging multimodal foundation models and self-distillation to further close this gap.
\section{Acknowledgement}
This research is funded in part by the National Natural Science Foundation of China (62171326, 62371350, 62471343), Key Science and Technology Research Project of Xinjiang Production and Construction Corps (2025AB029), Guangdong OPPO Mobile Telecommunications Corp. and Wuhan University Supercomputing Center.
{
    \small
    \bibliographystyle{ieeenat_fullname}
    \bibliography{main}
}


\end{document}